\documentclass[conference]{IEEEtran}
\usepackage{graphicx, amsmath,subfigure, multirow, array, amsfonts,paralist,
algorithmic, algorithm, color}

\begin{document}

\title{Behavioral Learning of Aircraft Landing Sequencing Using
a Society of Probabilistic Finite State Machines}

\author{\IEEEauthorblockN{Jiangjun Tang\IEEEauthorrefmark{1} and
Hussein A. Abbass\IEEEauthorrefmark{2} } \IEEEauthorblockA{School
of Engineering and Information Technology, UNSW-Canberra,
Canberra, Australia\\ \IEEEauthorrefmark{1}Email:
j.tang@adfa.edu.au} \IEEEauthorblockA{\IEEEauthorrefmark{2}Email:
h.abbass@adfa.edu.au}}

\maketitle

\makeatletter
\def\ps@IEEEtitlepagestyle{
  \def\@oddfoot{\mycopyrightnotice}
  \def\@evenfoot{}
}
\def\mycopyrightnotice{
  {\footnotesize
  \begin{minipage}{\textwidth}
  \centering
  Copyright~\copyright~2014 IEEE. J. Tang and H. A. Abbass, "Behavioral learning of aircraft landing sequencing using a society of Probabilistic Finite state Machines," 2014 IEEE Congress on Evolutionary Computation (CEC), Beijing, 2014, pp. 610-617.
doi: 10.1109/CEC.2014.6900597
  \end{minipage}
  }
}


\begin{abstract}

Air Traffic Control (ATC) is a complex safety critical
environment. A tower controller would be making many decisions in
real-time to sequence aircraft. While some optimization tools
exist to help the controller in some airports, even in these
situations, the real sequence of the aircraft adopted by the
controller is significantly different from the one proposed by the
optimization algorithm. This is due to the very dynamic nature of
the environment.

The objective of this paper is to test the hypothesis that one can
learn from the sequence adopted by the controller some strategies
that can act as heuristics in decision support tools for aircraft
sequencing. This aim is tested in this paper by attempting to
learn sequences generated from a well-known sequencing method that
is being used in the real world.

The approach relies on a genetic algorithm (GA)  to learn these
sequences using a society Probabilistic Finite-state Machines
(PFSMs). Each PFSM learns a different sub-space; thus, decomposing
the learning problem into a group of agents that need to work
together to learn the overall problem. Three sequence metrics
(Levenshtein, Hamming and Position distances) are compared as the
fitness functions in GA. As the results suggest, it is possible to
learn the behavior of the algorithm/heuristic that generated the
original sequence from very limited information.

\end{abstract}

\section{Introduction}
To find optimal landing sequences for arrival aircraft is an
NP-hard problem when the constraints of spacing between arrivals
depend on aircraft types and other
conditions~\cite{hu2005genetic}. The First-Come-First-Served
(FCFS) heuristic has been used as the most common approach for
sequencing aircraft~\cite{odoni1994models}. FCFS simply schedules
the landing aircraft based on the Estimated Time of Arrival (ETA)
at the runway and the minimum separation time between two
consecutive aircraft as listed in Table~\ref{Tab:MinSeparation}.

\begin{table}[h]
  \centering
  \begin{tabular}{|c|c|c|c|} \hline
    \multirow{2}{*}{Leading Aircraft} & \multicolumn{3}{c|}{ Trailing
    Aircraft}\\\cline{2-4}
    & Heavy & Large & Small\\ \hline
    Heavy & 96 & 157 & 196\\ \hline
    Large & 60 & 69 & 131\\ \hline
    Small & 60 & 69 & 82\\      \hline
  \end{tabular}
   \caption{Minimum time separation (in seconds)
   between landings and mandated by FAA~\cite{de2002airport}}\label{Tab:MinSeparation}
\end{table}

FCFS schedule is easy to be implemented, and it maintains a sense
of fairness. However, the landing efficiency in terms of runway
throughput can't be guaranteed by FCFS when unnecessary spacing
requirements exist. Therefore, many aircraft landing sequencing
approaches have been proposed and automation tools have been
deployed in the operational environment to increase the efficiency
of the system by maximizing runway throughput while maintaining
safety.

Constrain Position Shifting
(CPS)~\cite{balakrishnan2006scheduling} is a common approach in
practice, which shifts an aircraft forward or backward in the FCFS
schedule by a specified maximum number of positions in order to
achieve a landing sequence with the smallest time span. These
approaches show some advantages over FCFS, such as providing the
Air Traffic Controller (ATC) with additional flexibility and
decision support plots to predict landing times and
positions~\cite{neuman1991analysis}. However, they increase
ATC-Pilot communication and controller workload. Therefore, a
mixed approach combining FCFS and aircraft shifting methods can be
a better option in the operational environment for trading-off
efficiency and ATC-pilot communications. Sequencing of aircraft
landing by ATC is a more complex procedure in the real world. Many
factors can come into play including weather conditions, emergency
situations, and even the personality and experience of an
ATC~\cite{wickens1997flight}. Any mistake made for aircraft
landing sequencing can cause critical safety risks in aviation.

Our previous work looked at the risk assessment of aircraft
landing sequencing algorithms~\cite{zhao2010evolutionary}. Some
critical issues were recognized by approaches based on the
Computational Red Teaming (CRT)~\cite{abbass2011computational}
concept. The previous work was not concerned with learning the
behaviors that generated a sequence. CRT usually starts with no or
limited knowledge about the object to be challenged at the
beginning. The behavior of an object needs to be learned by CRT
through observations. In order to understand aircraft landing
sequencing behavior and then provide some assessments, a
methodology is needed to learn and model the behavior, which can
enable us to apply CRT to challenge it specifically and thus,
improving it.

In this paper, we present a stochastic approach combined with a
Probabilistic Finite-state Machine
(PFSM)~\cite{rabin1963probabilistic} and Genetic
Algorithm(GA)~\cite{goldberg1989genetic} to represent and learn
the behavior of aircraft landing sequencing. We developed a
simulator for simulating this type of behavior, where either FCFS
and CPS operate on aircraft landing sequencing in order to balance
runway efficiency and ATC-pilot communications. The current
simulator considers some traffic conditions including mixed
aircraft types and their ETA sequence.

A Finite-state Machine (FSM) has been widely used as a
representation in many domain of applications such as, behavioral
modelling in simulating autonomous entities~\cite{Donikian2001},
machine learning~\cite{mitchell1997machine}, and pattern
recognition~\cite{vidal2005probabilistic}. It has also been
applied with Evolutionary Computation for solving various problems
~\cite{fogel1993evolving}~\cite{fogel1994Introduction}~\cite{abbass2004learning}.
In our approach, PFSM is used to simulate the behavior for
sequencing landing aircraft and then GA improve it by evolving the
transition probabilities for generating aircraft landing sequences
with high similarity to the targeting landing sequencing from the
behavior simulator. In our approach, the similarity is measured by
three sequence metrics: Levenshtein~\cite{levenshtein1966binary},
Hamming~\cite{steane1996error} and Position based distances. GA is
applied to each of them to evolve PFSM. A comparison among the
metrics is conducted for finding a better measurement.

This paper is organized as follows. The behavior simulator and
problem definition of aircraft landing sequencing are described in
Section~\ref{Sec:Simulator}. This is followed by the proposed
methodology and the three metrics for learning and modelling the
behavior. Finally, the experimental results are presented for both
training and testing.

\section{Problem Definition and Aircraft Landing Sequencing
simulator}\label{Sec:Simulator}

Given aircraft arrival sequence $A(a_1, a_0, \ldots, a_n)$ and
their corresponding wake turbulence $W(w_1, w_2, \ldots,w_n)$,
aircraft landing sequence $A'(a_1', a_0', \ldots, a_n')$ can be
scheduled by some approach in order to maximize/minimize certain
objectives, e.g. increasing runway throughput and maintaining
safety. The wake turbulence of aircraft is classified into three
catalogs: Heavy ($H$), Light ($L$), and Small ($S$). The minimum
separation is defined in Table~\ref{Tab:MinSeparation} and is
required to constrain the inter-landing time of two subsequent
aircraft.

According to the minimum separation requirements and given
sequences ($A$) and wake turbulence ($W$), CPS can search for and
construct a sequence with the minimum time span for the landing of
all aircraft. Sometimes the re-scheduled aircraft landing sequence
($A'$) is not necessarily better than the FCFS sequence
($A$)~\cite{zhao2010evolutionary}. The data used in the simulator
represent peak time data; that is, the intervals between every two
aircraft estimated time of arrival (ETA) is always less than 1
minute. This inter-arrival time constraints the schedule because
it is mostly less than the minimum separation requirement. 1-CPS
is used in the simulator, which means aircraft position can be
shifted forward or backward by 1 position only. If a time span of
an aircraft landing sequence ($A'$) from CPS is better than the
FCFS sequence ($A$) by a certain level, the CPS sequence ($A'$) is
used otherwise the FCFS sequence ($A$) is used.

In summary, the aircraft landing sequencing simulator takes a
number of aircraft arrival sequence ($A_1,A_2,\ldots, A_m$) with
associated wake turbulence ($W_1, W_2, \ldots, W_m$) and generates
a number of new aircraft landing sequencing ($A'_1,A'_2,\ldots,
A'_m$) using a combination of CPS and FCFS actions in order to
balance efficiency and ATC-pilot communications.

The objective of this paper is to present an approach for learning
and modelling the behavior of this simulator using a Probabilistic
Finite-state Machine (PFSM) and Genetic algorithm (GA). The
methodology assumes knowledge of the following information:
\begin{itemize}
  \item the original arrival aircraft sequences($A_1,A_2,\ldots, A_m$) and their
  corresponding wake turbulence ($W_1, W_2, \ldots, W_m$), and
  \item a new sub-sequences generated by either CPS or FCFS. These optimized
  sub-sequences are used to train the algorithm.
  (training set).
\end{itemize}
However, the algorithm doesn't know any of the following
information:
\begin{itemize}
  \item The intents of the behavior; that is, whether the objective is to minimize or maximize the time span for landing
  all arrival aircraft,
  \item the underlying mechanism/algorithm being used for sequencing aircraft landing,
  \item the threshold that was used to select either CPS or FCFS, and
  \item the minimum separation requirements for different wake turbulence.
\end{itemize}

\section{Methodology}\label{Sec:Method}

\subsection{Probabilistic Finite-state Machine}\label{Sec:FSM}
A Probabilistic Finite-state Machine (PFSM) can be defined as a tuple
$A = (Q_A,\sum, \delta_A, I_A, F_A, P_A)$ ~\cite{vidal2005probabilistic}, where
\begin{itemize}
  \item $Q_A$ is a finite set of states;
  \item $\sum$ is a finite input alphabet;
  \item $\delta_A \subseteq Q_A \times \sum \times Q_A$ is a set of transition;
  \item $I_A:Q_A \rightarrow \mathbb{R}^+$ is the initial-state probabilities;
  \item $P_A:\delta_A \rightarrow \mathbb{R}^+$ is the transition probabilities;
  \item $F_A:Q_A\rightarrow \mathbb{R}^+$ is a set of acceptable states.
\end{itemize}
$I_A$, $P_A$ and $F_A$ are functions as below:
\begin{center}
$\sum_{q\in Q_A}I_A(q) = 1$
\end{center}
and
\begin{center}
$\forall q \in Q_A$, $F_A(q) + \sum_{a\in \sum, q'\in Q_A} P_A(q,a,q') =1$
\end{center}

Our approach manipulates the aircraft landing sequence according
to only two known inputs: ($A_1,A_2,\ldots, A_m$) and their wake
turbulence ($W_1, W_2, \ldots, W_m$). Here, the sequence of mixed
wake turbulence is the only meaningful information of traffic
conditions. In order to utilize it, a PFSM can be constructed by
taking each combination of wake turbulence as a state($Q$), whose
immediate next states are all permutations of this wake turbulence
combination. The transitions $\delta$ between them are governed by
a set of given probabilities($P$). However, the total number of
states will be $3^n$; given that there are three wake turbulence
types and $n$ aircraft in a sequence. The computation cost for
evaluating and evolving such kind of PFSM increases exponentially
when the number of aircraft increasing.

\begin{figure*}[t]
    \center
    \subfigure[PFSM1: three aircraft with three wake turbulence (H,L, and S)]{
        \includegraphics[width=0.48\textwidth]{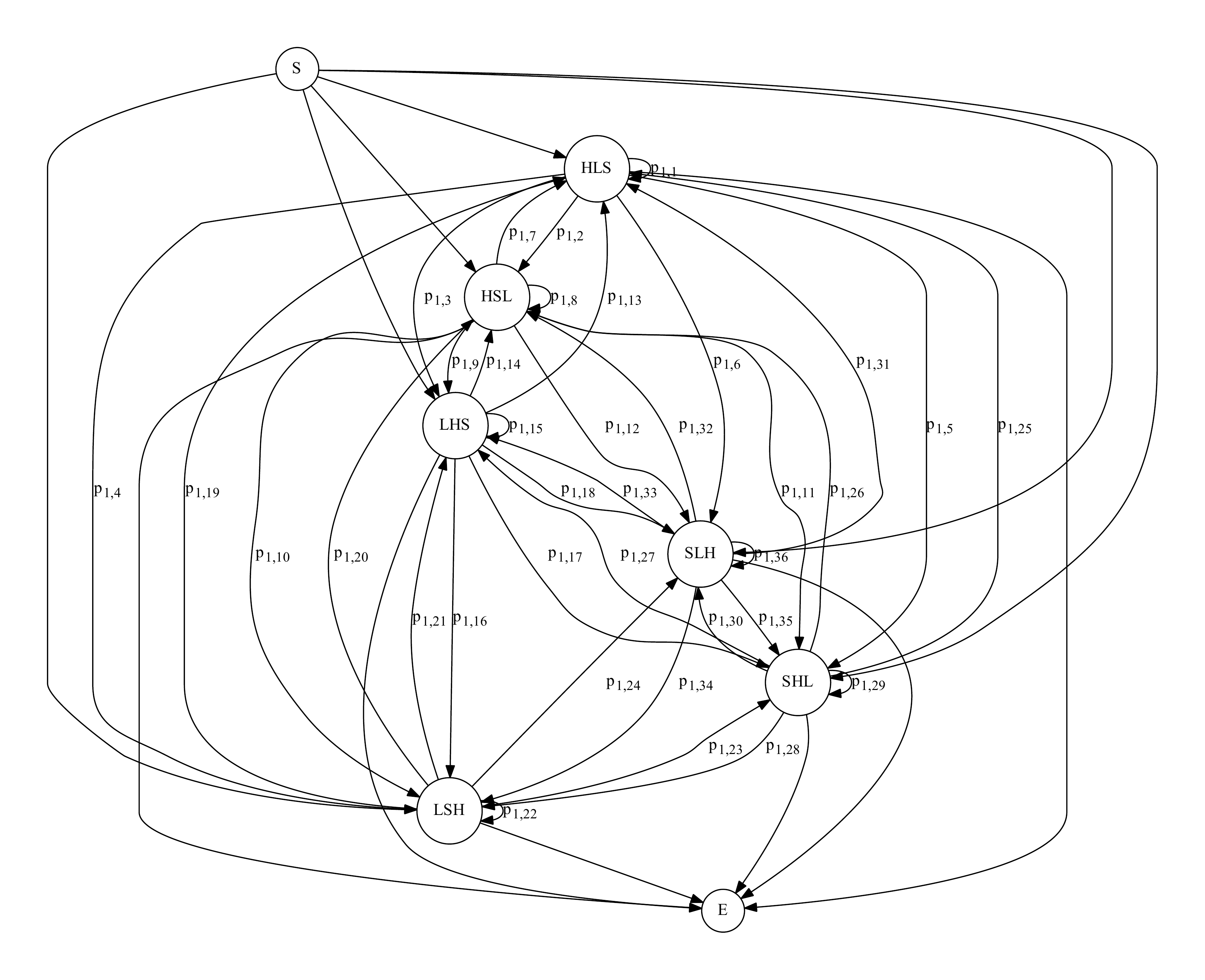}
        \label{fig:PMSM_three}
    }
    \subfigure[PFSM2: three aircraft with two wake turbulence (H and L)]{
        \includegraphics[width=0.48\textwidth]{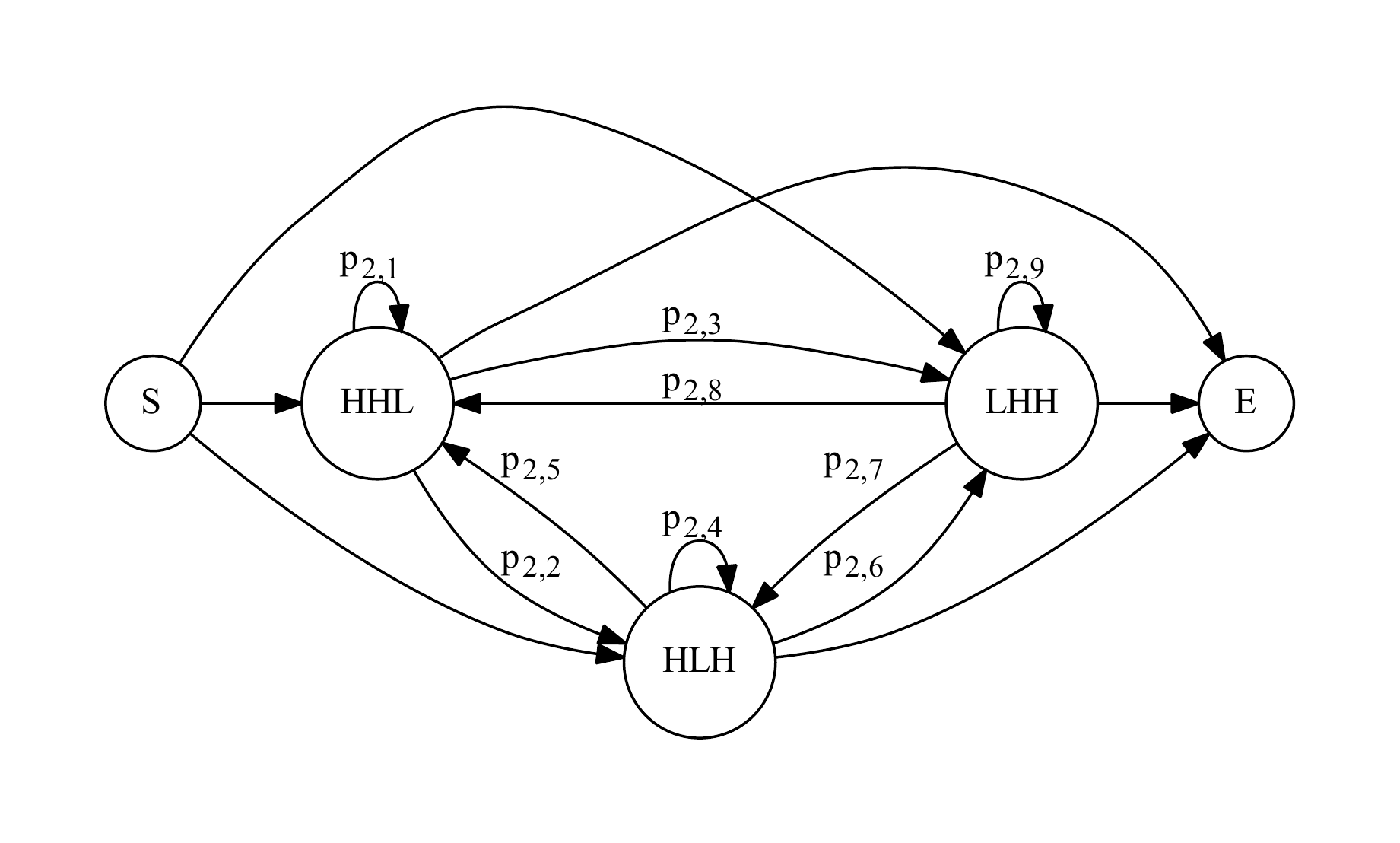}
        \label{fig:PFSM_two}
    }
    \caption{Two examples of PFSM for scheduling a sub-sequence of landing
       aircraft with different wake turbulence (H: Heavy, L: Large, and S:
       Small). $p_{i,j}$ is the transition probabilities, $i$
       is the index of a PFSM, and $j$ is the index of a
       probability.}\label{fig:PFSM}
\end{figure*}

Instead of building a very complex PFSM containing all
permutations as states, we construct a number of smaller PFSMs.
Each of them contains a group of related permutations (as states)
containing the same wake turbulence types. Therefore, we decompose
the aircraft sequence into a set of sub-sequences with length of
3. We use this short length to capture the number of wake
turbulence types. There is a total of 27 combinations that can be
generated, but three of them (``HHH'', ``LLL'', ``SSS'') is not
necessary to be included into the PFSM. This is mainly because
regardless of the sequencing algorithm being used, the wake
turbulence constraint is inactive for these sequences. Hence, a
total of 7 PFSMs are constructed as listed in
Table~\ref{Tab:PFSM}.

\begin{table}[h]
  \centering
  \begin{tabular}{|l|l|r|} \hline
    PFSM  & States (permutations) & Number of Transitions\\ \hline
    PFSM1  & HLS, HSL, LHS, LSH, SHL, SLH & 36\\ \hline
    PFSM2  & HHL, LHH, HLH  & 9\\ \hline
    PFSM3  & HHS, SHH, HSH  & 9\\ \hline
    PFSM4  & LLH, HLL, LHL  & 9\\   \hline
    PFSM5  & LLS, SLL, LSL  & 9\\   \hline
    PFSM6  & SSH, HSS, SHS  & 9\\   \hline
    PFSM7  & SSL, LSS, SLS  & 9\\   \hline
  \end{tabular}
   \caption{Probabilistic Finite-state Machines based on wake
   tuburlance}\label{Tab:PFSM}
\end{table}

Figure~\ref{fig:PFSM} depicts two examples of the PFSMs (PFSM1 and
PFSM2). Figure~\ref{fig:PMSM_three} presents a PFSM for an
aircraft sequence with three aircraft and each of them belongs to
one wake turbulence. Another example of three aircraft with only
two turbulence types H and L is presented in
Figure~\ref{fig:PFSM_two}. The structure of other PFSMs with only
two aircraft turbulence types is similar.

As shown in the figures, There are no probabilities associated
with the transitions between an initial state and its next states
or between a state and the end state. In effect, this means that
there are multiple initial states and multiple end states. The use
of a single start and end state in the representation is mainly
for convenience.


The different proposed PFSMs work together on a given arrival
aircraft sequence to generate an aircraft landing sequence by
Algorithm~\ref{Alg:PFSM}. The parameter $s$ defines the number of
aircraft in a temporary sequence that can be saved into the final
aircraft landing sequence.

\begin{algorithm}[h]
\caption{Aircraft landing sequencing by PFSM}
\label{Alg:PFSM}
\scriptsize
\begin{algorithmic}[1]
    \STATE \COMMENT {Input: arrival aircraft sequence $(A(a_1,a_2,\ldots,a_n))$
    with wake turbulence $(W(w_1,w_2,\ldots, w_n))$}
    \STATE Initial a empty sequence $A'' $
    \STATE Create two empty lists of $TempA$ and $TempW$
    \STATE Initial a integer step $s$, ($\in [1,3]$)
    \STATE Get the first three aircraft $(a_1, a_2, a_3)$ and their
    $(w_1,w_2,w_3)$
    \IF {$w_1 \equiv w_2 \equiv w_3$}
        \STATE Put $(a_1, a_2, a_3)$ into a new sequence $S$
    \ELSE
        \STATE Select a $FPSM_k$ has a state of $(w_1, w_2, w_3)$
        \STATE $FPSM_k$ generate a new sequence of $S$
    \ENDIF
    \STATE Put the first $s$ aircraft of $S$ into $A''$
    \STATE Put the rest aircraft into $TempA$ and the associated wake turbulence
    into $TempW$
    \FOR {i = 3 \bf{to} $n$ \bf{step} 1}
        \IF{length of $TempA$ ($l$) $\equiv 3$}
            \IF {all wake turbulenc in $TempW$ are same}
                \STATE Put $TempA$ into a new sequence of $S$
            \ELSE
                \STATE Select a $FPSM_k$ has a state of $TempW$
                \STATE $FPSM_k$ generate a new sequence of $S$
        \ENDIF
        \STATE Empty $TempA$ and $TempW$
        \STATE Put the first $s$ aircraft of $S$ into $A''$
        \STATE Put the rest aircraft of $S$ into $TempA$ and the associated
            wake turbulence into $TempW$
        \ELSE
            \STATE Add $a_i$ to the end of $TempA$ and $w_i$ to the end of $TempW$
        \ENDIF
    \ENDFOR
    \IF{length of $TempA$ ($l$) $> 0$}
        \STATE Move the last ($3-l$) aircraft from $A''$ and add them into $TempA$
        \STATE Add the related wake turbulence into $TempW$
        \STATE Select a $FPSM_k$ has a state of $TempW$
        \STATE $FPSM_k$ generate a new sequence of $S$
        \STATE Put the whole sequence $S$ at the end of $A''$
    \ENDIF
\end{algorithmic}
\end{algorithm}

Let us take an example. Assume a sequence of ``HSLH'', the first
three aircraft ``HSL" will be input to PFSM1 and generate a new
sequence. Assume that the output from PFSM1 is $SHL$ because the
transition probability, $p_{1,11}$, between $HSL$ and $SHL$ is the
largest transition probability in the learnt model ({\it i.e. a
maximum likelihood approach}). If the step ($s$) is defined as 1,
then the first aircraft ($S$) is pushed into a new sequence
($A''$), and the rest of the aircraft $HL$ and the forth aircraft
($H$) in arrival sequence form a new temporary sub-sequence
``HLH''. Therefore, PFSM2 is selected. A final output is one of
``HLH'', ``HHL'', and ``LHH'' based on the given transition
probabilities ($p_{2,4}$, $p_{2,5}$, and $p_{2,6}$). Let us
suppose that state ``HHL'' is selected as the output of the second
PFSM, and let us assume that this is the last sub-sequence, this
complete sub-sequence is pushed to the end of $A''$. In this way,
a new aircraft landing sequence ($SHHL$) is constructed.

As demonstrated by the example, the proposed PFSMs and algorithm
is capable of re-constructing aircraft landing sequencing based
only on the arrival sequence of aircraft and their wake turbulence
without any knowledge of the actual decision module or algorithm
being used to generate such sequence.

Since this is a probabilistic approach, it is necessary to develop
an algorithm to learn the probabilities and as such, guides the
PFMS to learn the real decision module that was used to generate
that sequence. Before we discuss this algorithm, we need to
discuss how two sequences are compared to calculate a measure of
merit or a similarity measure.

\subsection{Sequence Metrics}\label{sec:Metrics}
As the aircraft sequence can be transferred into a string sequence
of aircraft types, string metrics are a suitable measurement to
evaluate the performance of the proposed PFSM when learning and
modelling of arrival aircraft sequencing behavior. Here, we use
three string metrics with different biases as described below.

\subsubsection{Levenshtein Distance}
Levenshtein distance~\cite{levenshtein1966binary} measures the
difference between two strings by the minimum number of single
character edits including insertions, deletions or substitutions
required to change one string into another. For two given aircraft
sequences ``SLLHLH'' and ``LLHLHS'',  the Levenshtein distance of
them is 2, which includes deleting the first 'L' and inserting an
'S' at the end of the second sequence. Therefore, the local
optimal alignment between two sequences is considered in
Levenshtein distance.

\subsubsection{Hamming Distance}
Hamming distance~\cite{steane1996error} is a widely used metric to
compare two strings with equal length. In our case, both original
arrival sequence and shifted sequence have equal length,
therefore, it is a suitable measurement for our PFSM. The Hamming
distance is the minimum substitutions required to change one
string to another; in other words, it is the number of mismatched
characters between two strings. It is classically used for binary
domains, but it is a generic metric independent of the size of the
alphabet set. For example, given two aircraft sequences of
``SLLHLH'' and ``LLHLHS'', the Hamming distance is 5.

\subsubsection{Position based Distance}
The third distance used in this paper is defined as the sum of the
distances between the position of an aircraft in the original
position and the shifted position of the same aircraft in the
shifted position. Equation ~\ref{EQ:PosD} describes the
calculation of it.
\begin{equation}\label{EQ:PosD}
PosD = \sum_{i=1}^l|P_i-P_i'|
\end{equation}
where $l$ is the length of the sequence, $P_i$ is the original
position of aircraft $i$ in the original sequence, and $P_i'$ is
the shifted position in the shifted sequence. For the giving
example, the position distance  is 10 ($5+1+1+1+1+1$) for the two
sequences ``SLLHLH'' and ``LLHLHS''  because the first aircraft of
'S' is shifted to the last position ({\it i.e.} distance is 5),
then all other five aircraft are shifted to the left by a single
position. It is important to emphasize that because of the
redundancy in the alphabets, when a letter is checked, it will be
compared to the closest position it moved to. In other words, the
first ``L" in the first sequence could be the first, second or
fourth ``L" in the second sequence. We always assume it is the
closest encounter; as such, it is the first ``L" in the second
sequence.

The Position based Distance is the most strict metric to measure
the global similarity between two sequences regardless of their
local similarities. The Levenshtein Distance considers is more
local than the other two. The effect of different metrics on
evaluating our PFSM for learning the behavior and modelling is
investigated in Section~\ref{Sec:results}.

\subsection{PFSM Evolution}\label{Sec:GA}
In our approach, Genetic Algorithm (GA) is used to train our
proposed PFSM to learn the behavior of aircraft landing sequencing
by evolving the transition probabilities.

The length of each chromosome is equal to the total number of
transitions (90) in all PFSMs as described before. The locus in
the chromosome is associated to a certain transition probability
in a certain PFSM. Each gene is a real number with a lower
boundary of 0 and an upper boundary of 100. Each chromosome has 7
building blocks which are mapped to the proposed 7 PFSMs
respectively. The length of each building block depends on the
number of transitions in a PFMS and they are varied, e.g. The
length of the building block for PFSM1 is 36 while the length of
the building block for others is 9. Each building block has
several sub-building-blocks associated with the transitions from
one state to others. To decode such a chromosome into the
transitions probabilities, a group of real numbers belonging to a
sub-building-block is converted into transition probabilities by
normalization.

For example, nine probabilities (from $p_{2,1}$ to $p_{2,9}$)
exist in PFSM2 as shown in Figure~\ref{fig:PFSM_two}. Therefore,
there are nine loci in a chromosome associated with this PFSM and
form one building block. This building block contains 3
sub-building blocks corresponding to the three groups of
transition probabilities: $(p_{2,1}, p_{2,2}, p_{2,3})$,
$(p_{2,4}, p_{2,5}, p_{2,6})$, and $(p_{2,7}, p_{2,8}, p_{2,9})$.
According to the definition of PFSM described in
Section~\ref{Sec:FSM}, the genes ($g$) on these loci are
normalized into transition probabilities satisfying:
\begin{center}
    $p_{2,1} + p_{2,2} + p_{2,3} = 1$ \\
    $p_{2,3} + p_{2,5} + p_{2,6} = 1$ \\
    $p_{2,7} + p_{2,8} + p_{2,9} = 1$
\end{center}

After determining the transition probabilities of PFSMs, an
aircraft landing sequence ($A''$) is produced by a given aircraft
arrival sequence ($A$). Then $A''$ is compared against the
corresponding landing sequences $A'$ from the aircraft landing
sequencing simulator by one of the string metrics ($d$) as
explained in the last section. As our PFMS is a scholastic
approach for aircraft landing sequencing, it requires multiple
evaluations of each chromosome to approximate the fitness:
\begin{enumerate}
  \item Construct a PFSM by a chromosome
  \item For each pair of given $A_i$ and $A'_i$
    \begin{enumerate}
       \item Generate an aircraft landing sequence ($A_{i,j}''$) by PFSM with the
       input of $A_i$
       \item Calculate the sequence metric: $d_{i,j} = Dist(A''_{i,j}, A_i')$;
       \item Repeat Step $a$ and $b$ until a number ($T$) of evaluations
       researched;
       \item Get the mode ($D_i$) from all $d_{i,j}$;
    \end{enumerate}
  \item Get the mode ($D$) from all $D_i$ as the fitness of the chromosome.
\end{enumerate}

From the above steps, the fitness of a chromosome can be defined by
Equation~\ref{EQ:Fitness}.
\begin{equation}\label{EQ:Fitness}
    F=Mode_{i=1}^{n}(Mode_{j=1}^{T}(Dist(A''_{i,j}, A'_i))
\end{equation}
Where, $i$ is the index of a pair of aircraft arrival and landing
sequences in a given set, $n$ is the total number of aircraft
sequences in a given set, $j$ is the index of an evaluation for a
chromosome, and $T$ is the total number of evaluations on a
chromosome. The function of $Dist$ can be any one of the three
sequence metrics as mentioned above.

The objective of our GA is to minimize $F$. Binary tournament
selection is used to choose parents and then Uniform Crossover is
applied for producing offspring. When the mutation happens, a
random real number between 0 and 100 replaces the old gene. GA
stops when a predefined number of generations is reached.

\section{Experiment and Results}\label{Sec:results}

\begin{figure*}[ht]
\center
\subfigure[Levenshtein distance]{%
            \label{fig:Lev}
        \includegraphics[width=0.48\textwidth]{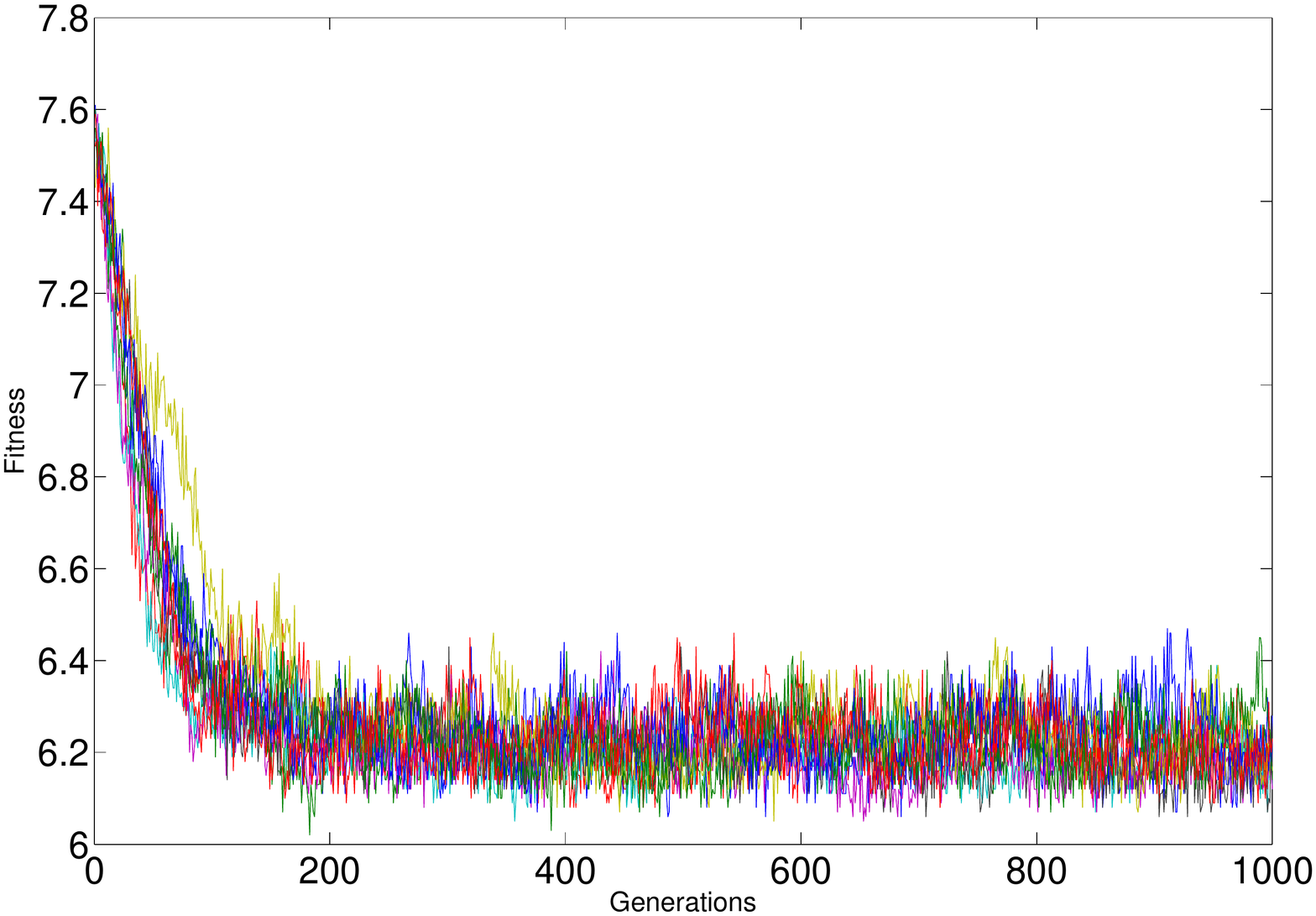}
        }%
        \subfigure[Hamming distance]{%
           \label{fig:Ham}
           \includegraphics[width=0.48\textwidth]{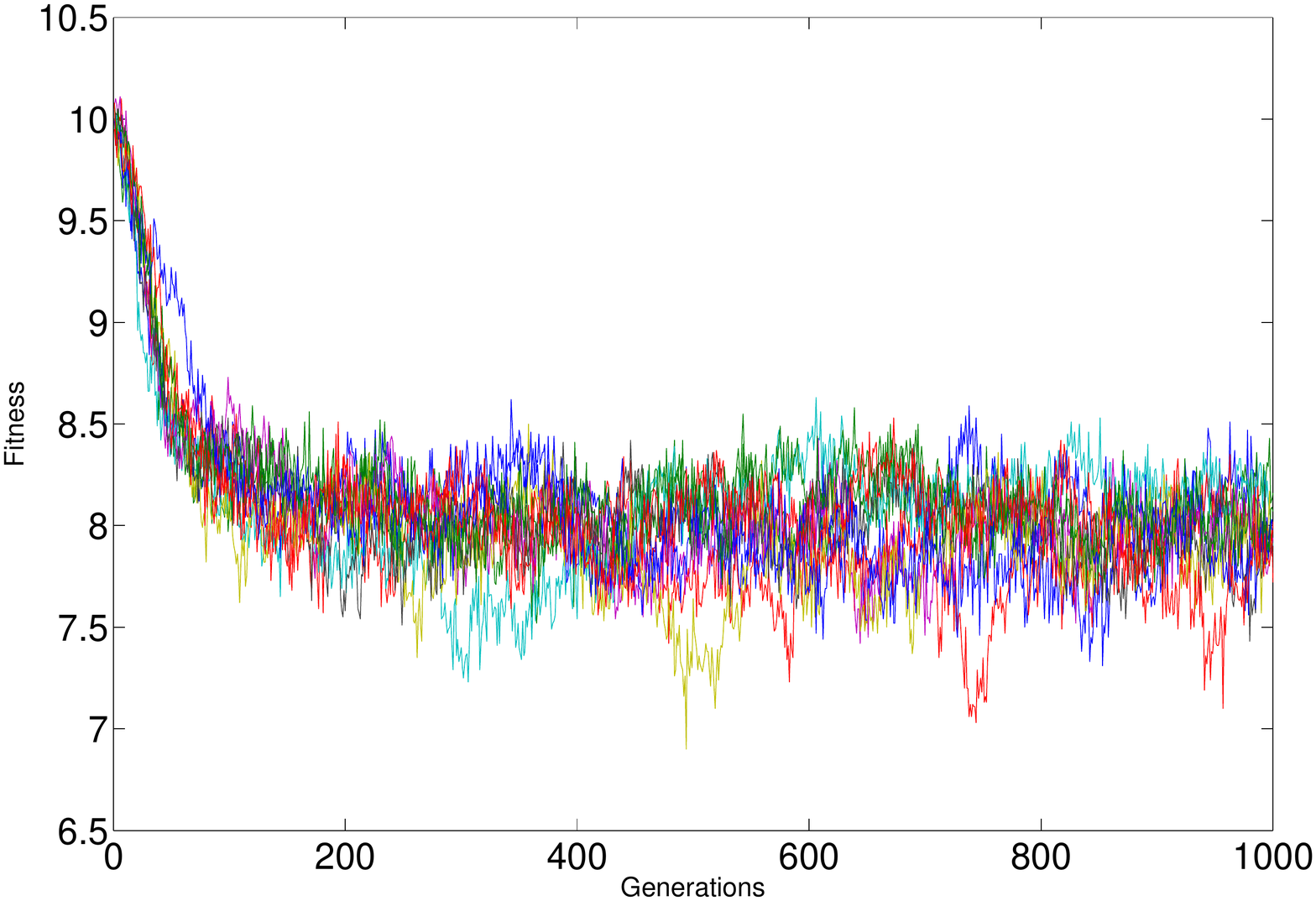}
        }\\
        \subfigure[Position distance]{%
           \label{fig:Pos}
           \includegraphics[width=0.48\textwidth]{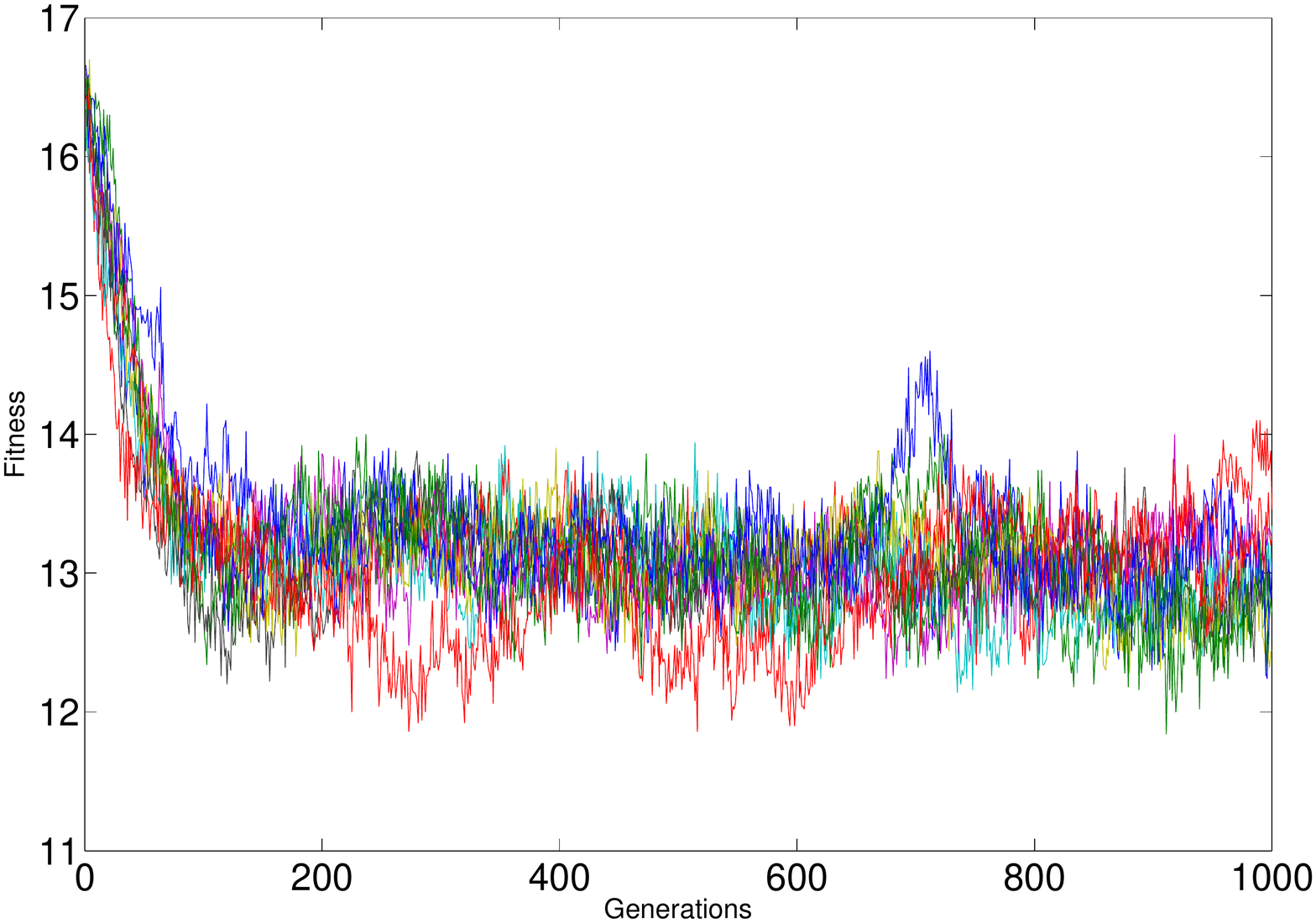}
        }
        \subfigure[Comparison of metrics]{%
           \label{fig:comp}
           \includegraphics[width=0.48\textwidth]{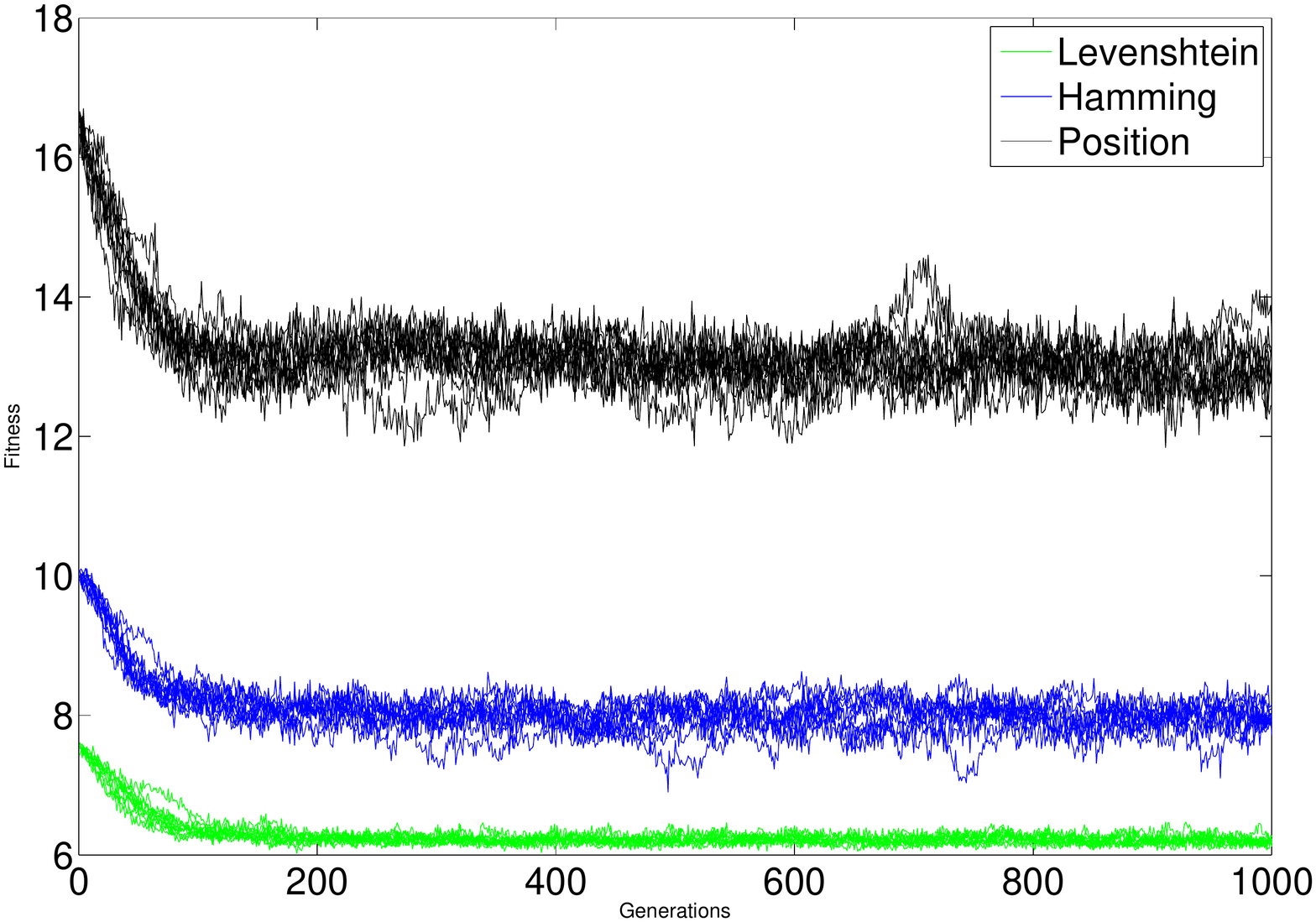}
        }
        \caption{The process in average fitness values for all three sequence
        metrics along with generations}\label{fig:GAGenerations}
\end{figure*}

\subsection{Experiment Design}
A total of 200 unique aircraft landing sequences (each sequence
has 20 aircraft) are generated randomly. They are all being fed
into the simulator. If the savings on the time span are at least
5\% time savings, the re-scheduled sequence from CPS is saved
otherwise FCFS is chosen. This is consistent with a realistic
operational environment constraints, where there is a need to
balance efficiency and ATC-pilot communications. The first half of
each sequence (100 aircraft) is used for training, while the
second half for testing.


A total of 90 transition probabilities is required to construct
all 7 PFSMs as explained in Section~\ref{Sec:FSM}.

GA is then used to evolve transition probabilities for training
the PFSM on the training set in order to learn and model the
aircraft landing sequencing behavior of the simulator. The
parameters used in GA is listed as follows:
\begin{itemize}
  \item population size: 100
  \item number of generation: 1000
  \item crossover rate: 0.9
  \item mutation rate: $1/l$, the reciprocal of the chromosome length ($l$),
  which is 90 in this experiment.
  \item The initial generation initializes the chromosomes randomly from uniform
distributions.
\end{itemize}

As we have three different sequence metrics for fitness
calculations, we run GA on each metric using ten different seeds.
The training results are presented in the next section.

After training, the best individual in the population is selected
from each of three metrics respectively and is tested on the test
set. The test results are provided in Section~\ref{Sec:Test}.

\subsection{Training Results}

The evolutions of our PFSM from three different sequence metrics
are shown in Figure~\ref{fig:GAGenerations}. The first three
illustrate the average fitness values along the generations for
three sequence metrics respectively. Figure~\ref{fig:comp} is
showing the comparison of different sequence metrics.

As expected, the fitness derived from Levenshtein distance has the
lowest magnitude and the fitness from Position distance has the
highest magnitude. The fitness from Position distance shows the
largest variations between runs as shown in Figure~\ref{fig:comp}.

\begin{figure}[h]
       \includegraphics[width=0.48\textwidth]{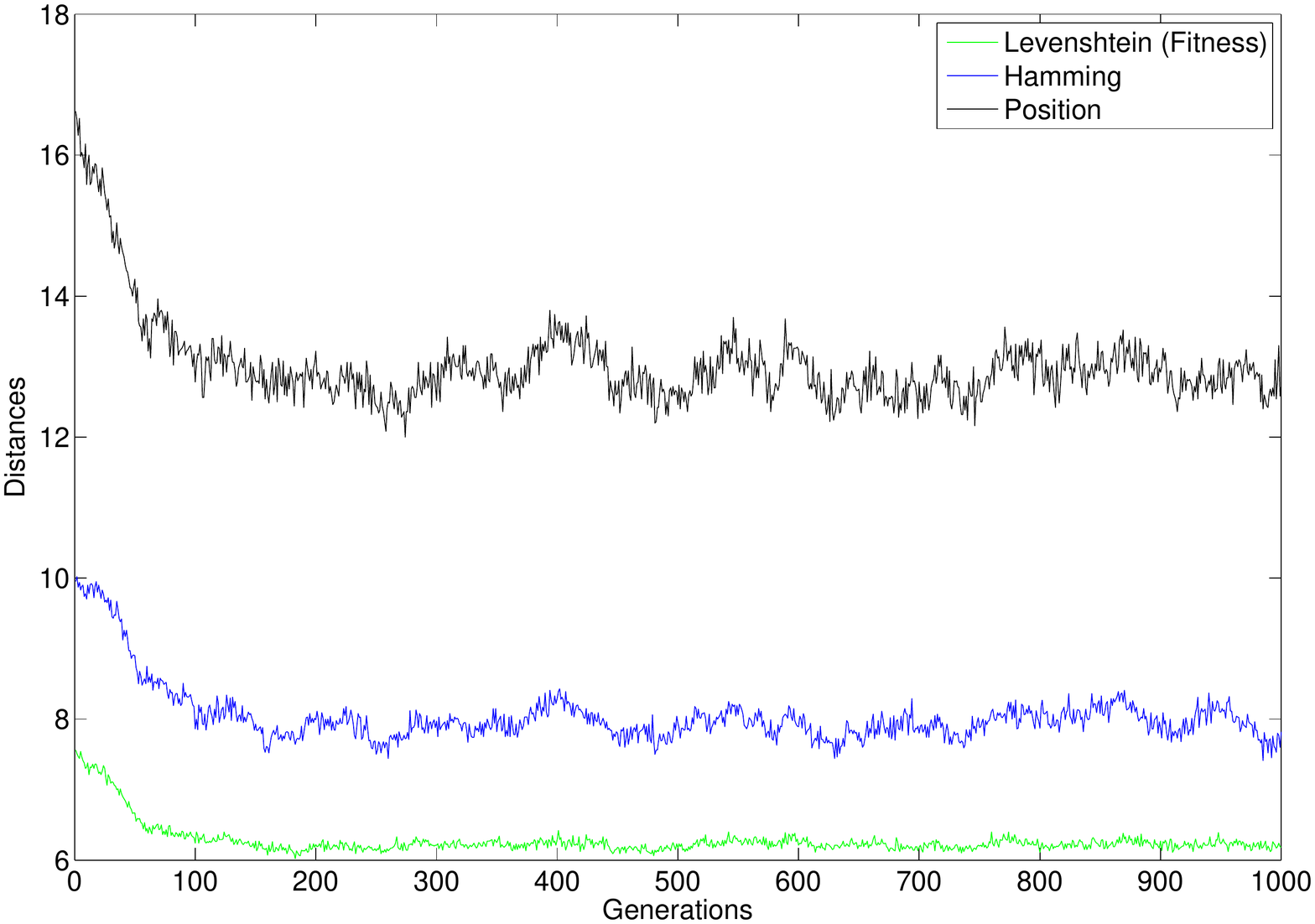}
       \caption{The best run of fitness calculated by Levenshtein distance and evaluated also on the other two metrics measured on the same
       individuals}\label{fig:BestLevInd}
\end{figure}

Since each metric is providing different magnitudes, it can happen
that the best solution found by each metric appears to be
different in fitness, but it is actually the same in terms of
decision variables ({\it i.e.} probabilities). Therefore, we
continue the analysis by taking the best solution found by one
metric in each generation and evaluate it also on the other two
metrics. An example of the best solution found over all runs using
Leveshtein distance is presented in Figure~\ref{fig:BestLevInd}.
This solution is also being evaluated on the other two metrics in
the figure.

Although the fitness is based on Levenshtein distance, the other
two generally follow the same trend of the fitness function. One
interesting point is that the fluctuations of Hamming distance is
quite similar to Position distance, although their magnitudes are
at different levels. Similar circumstances are also found when we
investigated the best runs from Hamming distance or Position
distance.

Since we are evaluating the same individual in the figure at each
generation, the fluctuations in the Levenshtein distance is due to
the stochastic nature of the solution. However, clearly there are
different sources causing different types of fluctuations when
this solution is evaluated on the other two metrics. To isolate
the two sources of fluctuations: those because of stochastic
representation and those because of the metric itself, we measured
the correlation coefficient between the Levenshtein distance and
other two. The correlation coefficient was 0.92 and 0.91 in
relation to the Hamming and Position distances respectively. This
indicates that there is a small amount of extra fluctuations that
are due to the metrics themselves.

Three best individuals in terms of their evaluated sequence
metrics from the 30 runs are selected and compared in
Table~\ref{Tab:BestInds}.
\begin{table}[h]
  \centering
  \begin{tabular}{|l|l|c|c|c|} \hline
    \multirow{2}{*}{Individuals} &\multirow{2}{*}{Fitness function} &
    \multicolumn{3}{c|}{ Sequence Metrics}\\\cline{3-5}
    & & Levenshtein & Hamming & Position\\\hline
    Ind\_Lev & Levenshtein &  \bf{4}& 6 & 10\\ \hline
    Ind\_Ham & Hamming  & 4 & \bf{4} & 12\\ \hline
    Ind\_Pos & Position  & 4 & 6 & \bf{6}\\     \hline
  \end{tabular}
   \caption{The fitness of the best individuals found by evolution using each metric and being evaluated on the other two sequence
   metrics.}\label{Tab:BestInds}
\end{table}

 \begin{figure*}[ht]
 \center
       \includegraphics[width=0.8\textwidth]{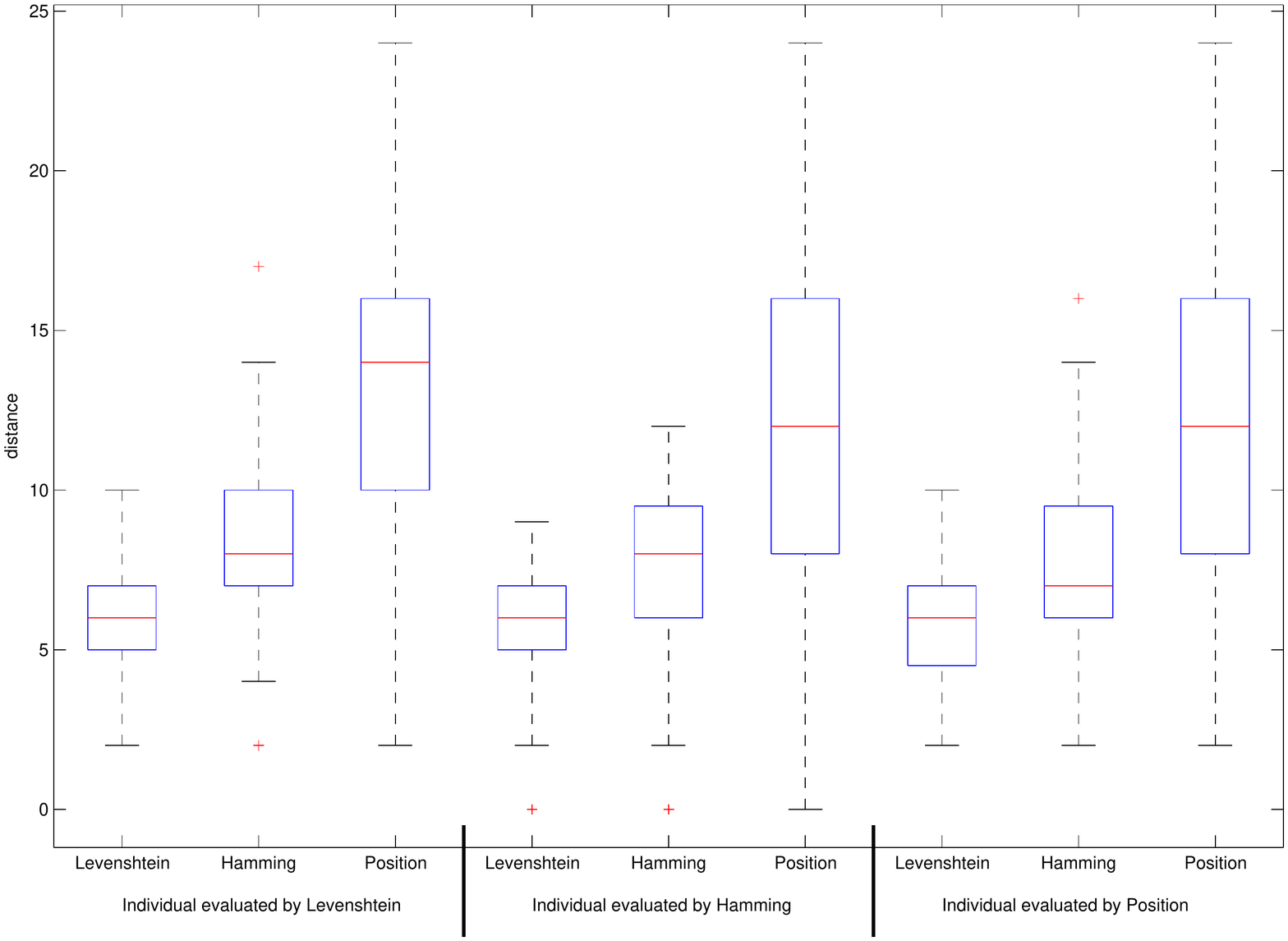}
       \caption{Whisker-box plots for the test results on three sequence
       metrics from all three best individuals obtained from
       training.}\label{fig:TestBox}
\end{figure*}

The individual evaluated by Levenstein distance has a small
Hamming distance but a larger the Position distance as shown in
the first row in the table. The individual evaluated by Position
distance is also able to produce both small Levenshein and Hamming
distances as listed in the last row. However, small Hamming
distance can't guarantee a small Position distance as being
demonstrated by the individual evaluated by Hamming distance in
the second row. These three individuals listed in the table are
also used for testing our approach using the test set in the next
section.

When looking at some specific aircraft landing sequencing, we find
that Levenshtein distance prefers to give low error in two
sequences when they have some local matching sequences. For
example, ``HHLSLLLSHHSLLSSSLHHL'' and ``HHSLLLLSSHHSLLSSLHHL'',
where Levenshtein distance is 4 because there are three local
sequence that are matched between two sequence, which are ``HH'',
``LLLS'', and ``SSLHHL''. However the Hamming distance is 6. But
the Position distance is 10 caused by the low global matching
between them.

\subsection{Test Results} \label{Sec:Test}

The aircraft arrival sequences in the test set is input to the
three individuals (PFSMs), Ind\_Lev, Ind\_Ham, and Ind\_Pos,
respectively. Each PFSM produces only one landing sequence for a
given arrival sequence using a maximum likelihood approach. The
new landing sequence is compared against the sequence from the
simulator. Table~\ref{Tab:BestIndTest} presents the mode of the
test results from three sequence metrics for all three
individuals. Similar to the training session, both individuals
that have been evaluated by the Hamming and Position distances can
produce small Levenshtein distance.

\begin{table}[h]
  \centering
  \begin{tabular}{|l|c|c|c|} \hline
    \multirow{2}{*}{Individuals} &
    \multicolumn{3}{c|}{ Sequence Metrics  (mode) }\\\cline{2-4}
    & Levenshtein & Hamming & Position\\\hline
    Ind\_Lev &  \bf{6}& 8 & 14\\ \hline
    Ind\_Ham &  7 & \bf{6} & 12\\ \hline
    Ind\_Pos & 4 & 6 & \bf{8}\\     \hline
  \end{tabular}
   \caption{The best individuals and their test results on three
   sequence metrics}\label{Tab:BestIndTest}
\end{table}

The distributions of distances on different sequence metrics
between the new sequences and the target sequence for each
individual are visualized as box charts in
Figure~\ref{fig:TestBox}. Overall, all the distance distributions
from these three individuals are quite similar to each other in
terms of means, 25th percentiles, and 75th percentiles. The
position distances produced by the three individuals all have
large variance but the variance of Levenshtein distance are always
smaller than the other two. As illustrated in the chart, the
individual evaluated by Hamming distance has the lowest outlier.

The number of produced sequences from each individual satisfying
the following conditions are counted and listed in
Table~\ref{Tab:NumSeq}:
\begin{itemize}
  \item the Levenshtein distance is less than 6 representing the mode
  of the Levenshtein distances of Ind\_Lev
  \item the Hamming distance is less than 6 representing the mode
  of the Hamming distances of Ind\_Ham
  \item the Position distance is less than 8 representing the mode
  of the Position distances of Ind\_Pos
\end{itemize}

\begin{table}[h]
  \centering
  \begin{tabular}{|l|c|c|c|} \hline
    \multirow{2}{*}{Individuals} &
    \multicolumn{3}{c|}{ Number of Sequences }\\\cline{2-4}
    & $Levenshtein \leq 6$ & $Hamming \leq 6$ & $Position \leq 8$\\\hline
    Ind\_Lev &  56 & 20 & 20\\ \hline
    Ind\_Ham &  64 & 37 & 31\\ \hline
    Ind\_Pos & 63 & 41 & 37\\   \hline
  \end{tabular}
   \caption{The number of sequences from each individual}\label{Tab:NumSeq}
\end{table}

As seen from the table, the individual evolved using Levenshtein
distance performed worse than the other two. The individual
evolved using Position distance has the best performance in terms
of high number of sequences below the given errors for all three
sequence metrics.

As demonstrated here, the PFSM approach is capable to learn and
model the aircraft landing sequencing behavior and produce good
matching landing sequences for the test data.

\section{Conclusion}\label{Sec:Conclusion}
In this paper, we proposed a stochastic approach combined with
PFSM and GA to learn and model aircraft landing sequencing
behavior. As the experiment results suggested, this approach is
capable to achieve the learning objective while only knowing
limited information.

Three different sequence metrics are used for the fitness function
in the GA. These metrics showed different preferences when
evolving the transition probabilities of PFSM. Levenshtein
distance considers the local matching sequence more than the other
two, while the Position distance is more strict in terms of global
matching. Therefore, the fitness values of Position distance shown
in the training session are more varied than the other two. All
three metrics are able to guide GA to find a set of good
transition probabilities for the proposed PFSM. However, the
global metrics, e.g. Position distance, has demonstrated the best
results on the test set, where it produced a higher number of
aircraft landing sequencing with lower errors than the other two.
The results show that the proposed approach is capable of learning
the underlying mechanism that generates a landing sequence.

In the future, we will introduce more uncertainty variables such
as weather conditions and emergency situations to disturb the
sequences during training. We will also investigate biased
initialization by relying on statistical estimation methods to
initialize the population in the GA instead of using a purely
random initialization approach. In general, learning the mechanism
that generated a solution can contribute to many sub-fields in
evolutionary computation including surrogate models,
simulation-based optimization, and fitness landscape analysis.

\section*{Acknowledgement }
This work has been funded by the Australian Research Council (ARC) discovery grant number, DP140102590: Challenging systems to discover vulnerabilities using computational red teaming.

\bibliographystyle{IEEEtran}

\end{document}